\documentclass{article}
\usepackage{iclr2027_conference,times}
\iclrfinalcopy
\usepackage[T1]{fontenc}
\usepackage[utf8]{inputenc}
\usepackage{amsmath,amssymb,booktabs,array,graphicx,xcolor,microtype}
\usepackage{float,hyperref,url}
\usepackage{tikz}
\usetikzlibrary{arrows.meta}
\usepackage{xcolor}
\usepackage{amsmath}
\definecolor{saclr}{RGB}{75, 110, 155}    
\definecolor{stclr}{RGB}{178, 122, 72}    
\hypersetup{hidelinks,pdftitle={Mathematical Transfer in LLMs Follows Reasoning Approach More Than Topic},pdfauthor={Anonymous authors}}
\newcommand{\SA}{\mathrm{SA}}
\newcommand{\ST}{\mathrm{ST}}

\title{Mathematical Transfer in LLMs Follows\\Reasoning Approach More Than Topic}

\author{Sajad Goudarzi\\
  Clemson University \\
  \texttt{sgoudar@clemson.edu}
\AND
Samaneh Zamanifard\\
  Clemson University\\
  \texttt{szamani@clemson.edu}
\AND
Seyed Amin Seyed Haeri\\
  Coastal Carolina University\\
  \texttt{sseyedha@coastal.edu}
\AND
 Moloud Nasiri\\
  Clemson University \\
\AND
Hamed Rahimian\\
  Clemson University \\
  \texttt{hrahimi@clemson.edu}
}

\begin{document}
\maketitle

\begin{abstract}
When selecting mathematical training data for LLMs, a natural organizing
principle is topic: probability examples for probability targets. An
alternative is reasoning approach: worked solutions that share a solution
method with the target, even when the mathematical domain differs. We ask
which relation produces greater transfer after fine-tuning. We evaluate
two counterbalanced $2\times2$ designs: probability and combinatorics
crossed with invariant reasoning and double counting (2{,}000 problems),
and number theory and geometry crossed with complement and pigeonhole
reasoning (800 problems). In each design, every cell serves as the held-out
target in turn: same-approach (SA) sources share the target's method but
change the topic, while same-topic (ST) sources share the topic but change
the method. Every source appears once in each role, so additive
source-quality effects cancel from the equally weighted aggregate
contrast. Across five base models and three training seeds per design,
SA outperforms ST in all 40 seed-pooled model--target comparisons.
Model-level advantages range from 8.2 to 16.2 percentage points in the
primary design (mean: 10.8) and from 12.0 to 16.0 in the second design
(mean: 14.3); all ten model-level 95\% confidence intervals exclude zero.
In both designs, ST sources are \emph{more} similar to targets under
embedding and lexical measures, so the SA advantage runs opposite to
the measured ordering of statement-level resemblance. These findings
identify reasoning approach as a more effective matching criterion than
topic for mathematical transfer across the evaluated topic--approach
combinations.
\end{abstract}

\section{Introduction}

What should mathematical training data share with the problems a
language model will be asked to solve? The most common organizing
principle is \emph{topic}: to improve performance on probability, train
on probability; to improve on combinatorics, train on combinatorics.
But mathematical problems can also be organized by \emph{reasoning
approach}, where a worked solution may exploit a conserved quantity
(invariant reasoning) or establish two counts of the same collection
(double counting), regardless of whether the objects are probabilistic
or combinatorial. These two axes can disagree: a source dataset can
match the target's topic while using a different solution method, or
match the method while changing the topic. Which relationship produces
greater transfer after fine-tuning?

Prior work gives reason to expect that approach matters. In human problem-solving, experts categorize problems by deep structural
principles rather than surface features, and structural correspondences are central to theories of analogical transfer
\citep{chi1981categorization,gick1983schema,gentner1983structure}. In
LLMs, skill-matched exemplars improve mathematical prompting over
topic-matched alternatives \citep{didolkar2024metacognitive}, and skill-aware data selection improves training efficiency
\citep{chen2023skill,he2026stat,zhang2026skill}. These findings
motivate our hypothesis, but leave a specific question open: after
\emph{parameter updates}, does supervision from a different topic that shares the target's approach outperform the same-topic supervision that
uses a different approach?

\begin{figure}[t]
\centering
\begin{tikzpicture}[
    cell/.style={
        minimum width=2.4cm, minimum height=1.2cm,
        draw=gray!40, line width=0.4pt,
        align=center, inner sep=3pt, font=\small},
    tcell/.style={cell, fill=black!8, draw=black!70, line width=0.8pt},
    sacell/.style={cell, fill=saclr!12, draw=saclr!45, line width=0.7pt},
    stcell/.style={cell, fill=stclr!12, draw=stclr!45, line width=0.7pt},
    ucell/.style={cell, fill=white, draw=gray!25},
    hdr/.style={font=\small\bfseries},
    dim/.style={font=\scriptsize\itshape, text=black},
    thumb/.style={minimum size=0.55cm, draw=gray!30, line width=0.3pt,
                  inner sep=0pt},
    tthumb/.style={thumb, fill=black!12, draw=black!50, line width=0.4pt},
    sathumb/.style={thumb, fill=saclr!22, draw=saclr!40},
    stthumb/.style={thumb, fill=stclr!22, draw=stclr!40},
    uthumb/.style={thumb, fill=white, draw=gray!20},
]
 
\def\cw{3.6}   
\def\rh{2.0}   
 
 
\node[font=\small\bfseries] at (-2.1, 1.75) {(a)};
 
\node[tcell]  (PI) at (0,0)
    {\textbf{P\,/\,I}\\[1pt]{\scriptsize TARGET}};
\node[sacell] (CI) at (\cw,0)
    {C\,/\,I\\[1pt]{\scriptsize\color{saclr!75} SA source}};
\node[stcell] (PD) at (0,-\rh)
    {P\,/\,D\\[1pt]{\scriptsize\color{stclr!70} ST source}};
\node[ucell]  (CD) at (\cw,-\rh)
    {\color{gray!40} C\,/\,D};
 
\node[hdr] at (0, 0.9)   {Probability};
\node[hdr] at (\cw, 0.9) {Combinatorics};
 
\node[hdr, anchor=east] at (-1.5, 0) {Invariant};
\node[hdr, anchor=east, align=right] at (-1.5, -\rh)
    {Double\\[-2pt]counting};
 
\node[dim] at (\cw/2, 1.35) {Topic};
\node[dim, rotate=90, anchor=south] at (-3.0, -\rh/2) {Approach};
 
\draw[-{Stealth[length=5pt, width=3.5pt]}, saclr!60, line width=1pt,
      shorten >=3pt, shorten <=3pt]
    (CI.west) -- (PI.east);
 
\draw[-{Stealth[length=5pt, width=3.5pt]}, stclr!60, line width=1pt,
      shorten >=3pt, shorten <=3pt]
    (PD.north) -- (PI.south);
 
\node[font=\scriptsize, anchor=north, align=left]
    at (\cw/2, -\rh - 0.7)
    {\textcolor{saclr!80}{\textbf{SA}} = Same Approach, different topic (same row)\\
     \textcolor{stclr!75}{\textbf{ST}} = Same Topic, different approach (same column)};
 
 
\def\ts{0.55}    
\def\tg{0.07}    
\pgfmathsetmacro{\cs}{\ts + \tg}  
 
\def\gax{6.6}   \def\gay{0.45}     
\def\gbx{8.9}   \def\gby{0.45}     
\def\gcx{6.6}   \def\gcy{-1.25}    
\def\gdx{8.9}   \def\gdy{-1.25}    
 
\node[font=\small\bfseries] at (\gax - 0.9, 1.75) {(b)};
\node[font=\scriptsize\itshape, text=black]
    at (7.75, 1.35) {All four target rotations};
 
\newcommand{\gridlabels}[2]{
    \node[font=\tiny, text=gray!45, anchor=south]
        at (#1, #2 + \ts/2 + 0.08) {P};
    \node[font=\tiny, text=gray!45, anchor=south]
        at (#1 + \cs, #2 + \ts/2 + 0.08) {C};
    \node[font=\tiny, text=gray!45, anchor=east]
        at (#1 - \ts/2 - 0.08, #2) {I};
    \node[font=\tiny, text=gray!45, anchor=east]
        at (#1 - \ts/2 - 0.08, #2 - \cs) {D};
}
 
\gridlabels{\gax}{\gay}
\gridlabels{\gbx}{\gby}
\gridlabels{\gcx}{\gcy}
\gridlabels{\gdx}{\gdy}
 
\node[tthumb]  at (\gax, \gay) {};
\node[sathumb] at (\gax + \cs, \gay) {};
\node[stthumb] at (\gax, \gay - \cs) {};
\node[uthumb]  at (\gax + \cs, \gay - \cs) {};
 
\node[sathumb] at (\gbx, \gby) {};
\node[tthumb]  at (\gbx + \cs, \gby) {};
\node[uthumb]  at (\gbx, \gby - \cs) {};
\node[stthumb] at (\gbx + \cs, \gby - \cs) {};
 
\node[stthumb] at (\gcx, \gcy) {};
\node[uthumb]  at (\gcx + \cs, \gcy) {};
\node[tthumb]  at (\gcx, \gcy - \cs) {};
\node[sathumb] at (\gcx + \cs, \gcy - \cs) {};
 
\node[uthumb]  at (\gdx, \gdy) {};
\node[stthumb] at (\gdx + \cs, \gdy) {};
\node[sathumb] at (\gdx, \gdy - \cs) {};
\node[tthumb]  at (\gdx + \cs, \gdy - \cs) {};
 
\def\ly{-2.55}
\def\lx{7.75}  
\node[fill=black!12, draw=black!50, line width=0.4pt,
      minimum size=0.3cm, inner sep=0pt] at (\lx - 1.5, \ly) {};
\node[font=\scriptsize, text=gray!55, anchor=west] at (\lx - 1.3, \ly) {Target};
\node[fill=saclr!22, draw=saclr!40, line width=0.3pt,
      minimum size=0.3cm, inner sep=0pt] at (\lx + 0.05, \ly) {};
\node[font=\scriptsize, text=gray!55, anchor=west] at (\lx + 0.25, \ly) {SA};
\node[fill=stclr!22, draw=stclr!40, line width=0.3pt,
      minimum size=0.3cm, inner sep=0pt] at (\lx + 1.0, \ly) {};
\node[font=\scriptsize, text=gray!55, anchor=west] at (\lx + 1.2, \ly) {ST};
 
\end{tikzpicture}
 
\vspace{-4pt}
\caption{\textbf{The $2\times 2$ rotational design}
\textbf{(a)}~Each cell pairs a topic (column) with a solution approach (row).
When one cell is held out as the target (here: probability/invariant), its SA source occupies the same row---shared approach, different
topic---while its ST source occupies the same column---shared topic,
different approach.
\textbf{(b)}~The target rotates through all four cells.}
\label{fig:design}
\end{figure}
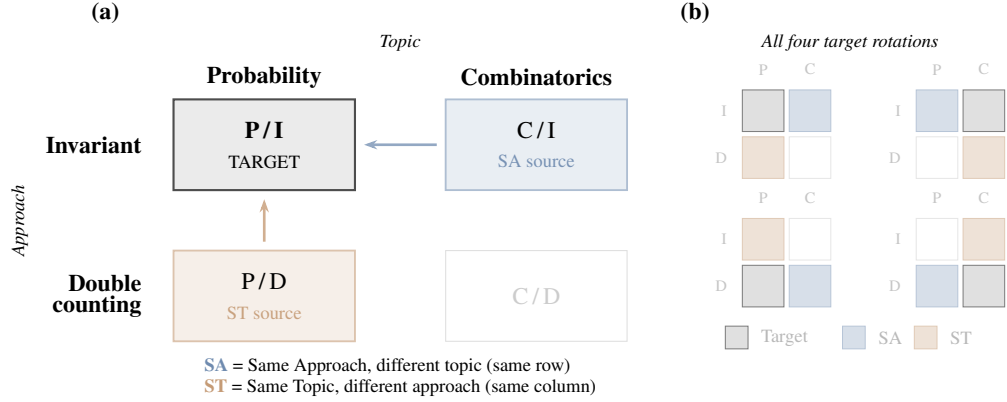

Answering this question requires a design that separates the
\emph{relationship} between source and target from the identity of the
source data. Comparing a single ``approach-matched dataset'' against a
single ``topic-matched dataset'' would confound the relationship with
the particular examples chosen. We therefore construct a $2\times2$
factorial crossing two topics (probability, combinatorics) with two
approaches (invariant, double counting) and rotate the held-out target
through all four cells (Figure~\ref{fig:design}). For each target, its
same-approach (SA) source shares the approach but changes the topic,
while its same-topic (ST) source shares the topic but changes the
approach. Because every source cell appears exactly once as SA and once as ST, 
additive source-quality differences cancel from the aggregate contrast.

Across five base models, three training seeds, and all four target
rotations, SA outperforms ST in every one of the 20 model--target
comparisons, with model-level advantages ranging from 8.2 to 16.2
percentage points (mean: 10.8). In 19 of 20 comparisons, SA also
improves over the unadapted base model, confirming that the advantage
reflects genuine positive transfer, not merely less harmful
adaptation. A statement-similarity analysis strengthens the finding:
same-topic problems are \emph{more} similar to targets under both
embedding and lexical measures, yet they produce weaker transfer.
The SA advantage, therefore, runs \emph{opposite} to the measured surface resemblance.

The ordering also replicates in an 800-problem design crossing number
theory and geometry with complement and pigeonhole reasoning. Across
the same five models and three seeds, all 20 seed-pooled model--target
contrasts again favor SA, extending the finding beyond the topic and
approach choices of the primary design.

Our contribution is a controlled, counterbalanced comparison of
topic-matched and approach-matched mathematical supervision,
replicated across five models, three seeds, and four target rotations.
The results identify the reasoning approach as a more effective axis than topic for selecting transfer-oriented fine-tuning data in the evaluated mathematical setting.

\section{Related work}
\label{sec:related}

A foundational finding in cognitive science is that experts and novices organize problems differently: experts classify physics problems by
underlying principles (e.g., conservation of energy), while novices
group them by surface features (e.g., inclined planes)
\citep{chi1981categorization}. This distinction extends to transfer. \citet{gick1983schema} show that analogical transfer depends on
abstracting a shared schema from surface-dissimilar examples, and
\citet{gentner1983structure} formalizes this as structure
mapping---the claim that transfer succeeds when relational structure,
not the surface objects, is preserved between the source and the target. These results motivate the hypothesis that the shared reasoning approach may
support mathematical transfer in LLMs more effectively than shared
topic. The cognitive evidence motivates the comparison, but does not
predetermine its outcome, since human categorization and learning from
gradient updates on worked solutions are different processes.

Within the LLM literature, the closest existing comparison is
\citet{didolkar2024metacognitive}, who derive mathematical skill
labels and find that skill-matched exemplars improve in-context
prompting over topic-matched alternatives. Our study differs in two respects: the intervention is fine-tuning (parameter updates) rather
than in-context demonstration, and the comparison uses a
counterbalanced rotation so that every source cell serves in both
roles. \citet{guo2024learning} further examines mathematical skill 
transfer under both in-context learning and instruction tuning. 
Importantly, while their in-context results show skill-sensitive effects, some instruction-tuning conditions produce similar changes
across skills, suggesting that transfer patterns observed in prompting
need not persist under parameter updates. This distinction motivates
our direct comparison of topic- and approach-matched transfer under source-specific fine-tuning.
Our factors are the mathematical topic and solution approach, and our primary endpoint is the paired
accuracy difference after source-specific fine-tuning.

More broadly, a growing body of work uses skill structure to guide
training-data selection: Skill-it \citep{chen2023skill} models
inter-skill dependencies to design training mixtures, LESS
\citep{xia2024less} selects influential examples via gradient-based
signals, STAT \citep{he2026stat} diagnoses missing skills and synthesizes targeted supervision, and skill-aware reasoning
distillation \citep{zhang2026skill} combines selection with explicit
skill decomposition. These methods ask \emph{how} to select training
data given a skill taxonomy. Our work addresses a complementary
question: \emph{which} axis of similarity---topic or reasoning
approach---produces greater transfer when the other is held out.

The form and content of mathematical supervision also affect
downstream performance more generally: chain-of-thought demonstrations
improve prompting \citep{wei2022cot}, and synthetically generated
questions support supervised adaptation \citep{yu2023metamath}. On the
evaluation side, benchmarks have become increasingly precise about the
kind of generalization they test. MATH \citep{hendrycks2021math}
provides competition-level problems with worked solutions; procedural
mathematics benchmarks \citep{saxton2019math} permit controlled
compositional splits; GSM-Symbolic \citep{mirzadeh2024gsm} probes
robustness to variation within fixed problem templates; and OMEGA
\citep{sun2025omega} distinguishes exploratory, compositional, and
transformative generalization. Our evaluation paradigm differs from
these: rather than measuring accuracy on an external benchmark, we
hold out one cell of a shared $2\times2$ construction pipeline and
compare how training on each remaining cell transfers to the held-out
target.

Finally, our design depends on distinguishing three properties that
need not coincide: final-answer correctness, solution-step validity,
and approach classification. \citet{lightman2024let} establish that
outcome-based and process-based verification target different aspects
of mathematical reasoning; this motivates our use of final-answer
accuracy as the primary transfer endpoint rather than
intermediate-step correctness. \citet{lee2026strategy} show that
lexical and embedding variation in solutions need not track variation
in underlying mathematical strategies; this motivates our treatment of
similarity measures as descriptive diagnostics rather than substitutes
for the approach rubric. The transfer endpoint accepts any valid
solution method; adherence to the reference approach is not required.

\section{A rotational test of mathematical transfer}
\label{sec:design}

\subsection{Cells and source--target relations}

Let $\mathcal T=\{P,C\}$ denote probability and combinatorics, and
$\mathcal A=\{I,D\}$ denote invariant reasoning and double-counting approaches.
The four cells are $\mathcal C=\mathcal T\times\mathcal A$. Each cell
contains problem statements, worked reference solutions, and final
answers. A cell label describes the intended organization of its
reference solutions; it does not preclude alternative valid methods
for the same problem.

At the conceptual level, invariant reasoning exploits a quantity or
relation preserved under specified transformations; probabilistic
formulations may use expectation-preserving quantities. Double
counting relates two enumerations of the same collection or incidence
structure. Boundary cases---including martingale arguments, potential functions,
mixed-method solutions, and sums of indicators---can blur the distinction
between the two approach categories.

For target $\tau=(t,a)$, write $\bar t$ and $\bar a$ for the
complementary topic and approach. The two comparison sources are
\begin{equation}
 s_{\SA}(\tau)=(\bar t,a),\qquad
 s_{\ST}(\tau)=(t,\bar a).
\label{eq:relations}
\end{equation}
A separate checkpoint is trained on each source cell. For a given
target, neither the SA nor the ST checkpoint has been trained on that
cell. Source and target roles rotate across the experiment; no
checkpoint is trained on the union of the four cells for the primary
comparison.

\subsection{Outcome and estimand}

Let $A_{m,s}(\sigma\!\rightarrow\!\tau)$ be final-answer accuracy, in
percent, on target $\tau$ for model $m$ and seed $s$ after training on
source $\sigma$. The target-specific contrast is
\begin{equation}
 \delta_{m,s,\tau}=A_{m,s}(s_{\SA}(\tau)\!\rightarrow\!\tau)
 -A_{m,s}(s_{\ST}(\tau)\!\rightarrow\!\tau).
\label{eq:delta}
\end{equation}
Our model-level estimand equally averages four targets and three seeds:
\begin{equation}
 \bar\delta_m=\frac{1}{12}\sum_{s\in\{7,17,27\}}
 \sum_{\tau\in\mathcal C}\delta_{m,s,\tau}.
\label{eq:estimand}
\end{equation}
For brevity, we refer to the first and second terms in \eqref{eq:delta} as $A_{\SA}$ and $A_{\ST}$, respectively. 
We also report $\Delta_{\SA}=A_{\SA}-A_{\rm base}$ and
$\Delta_{\ST}=A_{\ST}-A_{\rm base}$. The target-specific contrast can equivalently be written as
$\delta=\Delta_{\SA}-\Delta_{\ST}$. A positive contrast $\Delta_{\SA}-\Delta_{\ST}$ does not imply that both conditions improve
over the base model; SA can exceed ST even when one or both degrade
absolute accuracy. The five-model mean summarizes the evaluated suite
and is not a population-level estimate.

\subsection{What rotation controls}

Suppose a transfer score admits the decomposition
\begin{equation}
 A(\sigma\!\rightarrow\!\tau)=\mu+u_\sigma+v_\tau+w_{\sigma,\tau},
\label{eq:decomp}
\end{equation}
where $u_\sigma$ and $v_\tau$ are additive source and target effects, respectively.
The target term cancels within each SA--ST comparison. Each source
appears once under each relation, so the $u_\sigma$ terms cancel after
averaging over targets. Under the additive model, a source cell that is uniformly better, easier, or noisier cannot, on its own, produce the equally weighted aggregate contrast.
What rotation does not eliminate are
source--target interactions $w_{\sigma,\tau}$, including
statement-level resemblance under embeddings or TF--IDF.
Section~\ref{sec:similarity} reports those checks. Rotation controls
for fixed source identity, not for every leftover.

\section{Dataset and experimental protocol}
\label{sec:methods}
\subsection{Construction, answer auditing, and labels}

The dataset contains 2{,}000 problems, 500 in each cell. Candidates
were generated using gpt-oss-120b at high reasoning effort
\citep{openai2025oss}, conditioned on topic and approach and
accompanied by explicit reference-solution steps. Each candidate was
then re-solved eight times without an approach instruction by the generator, and three
large LLM judges, namely GLM-5.3, Deepseek-V4-Pro, and GPT5.6-terra, independently labeled the approaches of
the resulting solutions. A problem was retained only if all three
judges agreed on its approach label and a token-similarity criterion
was satisfied. This procedure selected problems compatible with the
generator's unconstrained behavior---problems whose natural solution
route matches the intended approach---rather than solutions forced to
follow a prescribed method. 

\begin{figure}[t]
\centering
\begin{tikzpicture}[
    stage/.style={
        draw=gray!50, line width=0.5pt, fill=gray!4,
        minimum width=3.3cm, minimum height=1.0cm,
        align=center, inner sep=4pt, font=\small,
        rounded corners=2pt},
    dataset/.style={
        draw=gray!50, line width=0.5pt, fill=gray!4,
        minimum width=5.5cm, minimum height=0.9cm,
        align=center, inner sep=5pt, font=\small,
        rounded corners=2pt},
    protocol/.style={
        draw=gray!50, line width=0.5pt, fill=gray!4,
        minimum width=11.0cm, minimum height=1.2cm,
        align=center, inner sep=6pt, font=\small,
        rounded corners=2pt},
    detail/.style={font=\scriptsize, text=black},
    connector/.style={
        -{Stealth[length=4pt, width=3pt]}, gray!40, line width=0.7pt},
    seclabel/.style={font=\scriptsize\bfseries, text=black},
]
 
\node[seclabel, anchor=west] at (-5.5, 3.3) {DATASET CONSTRUCTION};
 
\node[stage] (gen) at (-3.85, 2.5)
    {Generate\\candidates};
\node[detail, anchor=north, text width=3.4cm, align=center]
    at ([yshift=-4pt]gen.south)
    {gpt-oss-120b\\topic + approach prompt};
 
\node[stage] (verify) at (0, 2.5)
    {Verify approach\\labels};
\node[detail, anchor=north, text width=3.5cm, align=center]
    at ([yshift=-4pt]verify.south)
    {8 unconstrained re-solves\\3 LLM judges + similarity};
 
\node[stage] (audit) at (3.85, 2.5)
    {Audit final\\answers};
\node[detail, anchor=north, text width=3.4cm, align=center]
    at ([yshift=-4pt]audit.south)
    {LLM judges confirm final answer};
 
\draw[connector] (gen.east) -- (verify.west);
\draw[connector] (verify.east) -- (audit.west);
 
\node[dataset] (data) at (0, 0.1)
    {\textbf{Final dataset:} 2{,}000 problems, 500 per cell\\[-1pt]
     {\scriptsize 450 training + 50 validation in each source cell}};
 
\draw[connector] ([yshift=-2pt]0, 1.2) -- (data.north);
 
\node[seclabel, anchor=west] at (-5.5, -1.1) {EXPERIMENTAL PROTOCOL};
 
\node[protocol] (exp) at (0, -2.0)
    {For each held-out target: fine-tune separately on
     SA source and ST source (rank-16 LoRA, 450 examples)\\[1pt]
     Evaluate both checkpoints on target cell (500 problems)
     \quad$\longrightarrow$\quad
     $\delta = A_{\mathrm{SA}} - A_{\mathrm{ST}}$};
 
\draw[connector] (data.south) -- (exp.north);
 
\node[detail, anchor=north, text width=10cm, align=center]
    at (0, -2.85)
    {5 models $\times$ 3 seeds $\times$ 4 target rotations
     = 60 training jobs, 20 model--target contrasts};
 
\end{tikzpicture}
 
\vspace{-2pt}
\caption{\textbf{Dataset construction and experimental protocol.}
\emph{Top:} problems are generated with topic-and-approach conditioning,
verified via unconstrained re-solution and multi-judge agreement,
and corrected through final-answer auditing.
\emph{Middle:} the final dataset contains 500~problems per cell,
split 450/50 for training and validation.
\emph{Bottom:} for each of four target rotations, a separate checkpoint
is fine-tuned on the SA source and the ST source and evaluated on the
held-out target; the primary outcome is their accuracy
difference~$\delta$.}
\label{fig:pipeline}
\end{figure}
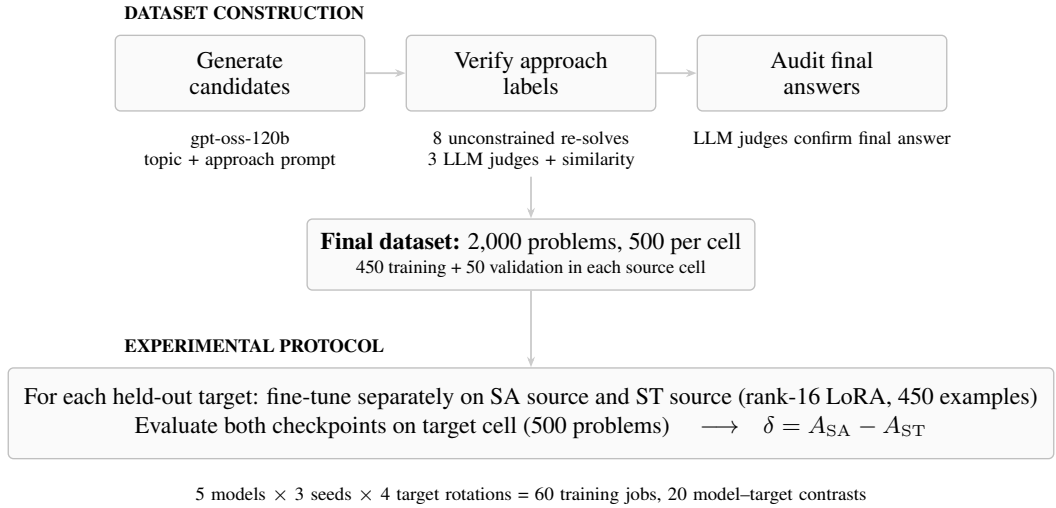

To verify the correctness of the final answers, GLM-5.3 and DeepSeek-V4-Pro were each asked to solve every problem twice without access to the reference solution. A problem was retained only when the solvers agreed on the answer. During further auditing, 83 of the 2{,}000 entries (4.15\%) were found to have inconsistent answers across the solvers; these entries were discarded and replaced using the same procedure. Finally, 50 randomly selected problems were manually reviewed, and all were found to be correct. All numerical results reported here use the final validated dataset.


\subsection{Models, training, and evaluation}
The primary suite comprises five base or pretrained checkpoints:
Qwen3-4B-Base \citep{qwen2025}, Meta-Llama-3-8B
\citep{meta2024llama}, DeepSeek-Math-7B-Base
\citep{shao2024deepseek}, Mistral-7B-v0.3 \citep{mistral2024v03},
and Gemma-3-4B-pt \citep{gemma2025}. This selection combines
general-purpose and math-specialized pretraining while remaining at
the base/pretrained level. Exact checkpoint identifiers are listed in
Appendix~\ref{app:protocol}.

Each model is adapted with rank-16 LoRA \citep{hu2021lora} on
reference solutions, using 450 examples from one source cell and
seeds 7, 17, and 27. The remaining 50 source-cell items serve as
validation. Cross-cell evaluation uses all 500 items of each held-out
target. A given item may appear in training for one rotational run
and in evaluation for another, but never in the checkpoint being
evaluated on its own cell.

This yields $5\times3\times4=60$ source-training jobs, all completed.
Each model--seed aggregate pools four target contrasts, producing 15
aggregates. Final-answer accuracy is the outcome; the evaluator does
not require the student to reproduce the reference wording or
approach.  

\subsection{Uncertainty and reporting}

Accuracy differences are reported in percentage points. We compute
95\% confidence intervals using paired, cell-stratified percentile
bootstraps over evaluation problems. For multi-seed estimates,
per-problem outcomes are first averaged across seeds before
resampling. The resulting intervals quantify uncertainty due to the
finite evaluation set conditional on the trained checkpoints; they do
not incorporate training-seed variability. Seed-to-seed variation is
therefore reported separately through the per-seed SA--ST contrasts.
Full details of the interval procedure are provided in
Appendix~\ref{app:statistics}.


\section{Results}
\label{sec:results}

\subsection{SA exceeds ST across all five models}

\begin{figure}[t]
\centering
\includegraphics[width=0.85\linewidth]{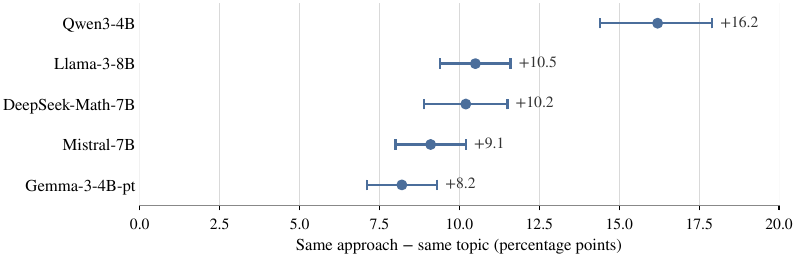}
\caption{\textbf{Approach-matched supervision transfers better in all
five models.} Each point averages four held-out targets and seeds 7,
17, and 27. Error bars are 95\% CIs; right labels show point
estimates. The mean across models is $+10.8$ percentage points.}
\label{fig:model}
\end{figure}

The model-level SA--ST advantages are $+16.2$ points for Qwen,
$+10.5$ for Llama, $+10.2$ for DeepSeek, $+9.1$ for Mistral, and
$+8.2$ for Gemma (Figure~\ref{fig:model}). Every reported 95\%
confidence interval lies above zero, and the equally weighted mean
across models is $+10.8$ points. All 15 model--seed aggregate
contrasts are also positive (Appendix~\ref{app:fullresults}).

The result is not driven by a few favorable comparisons. All 20
seed-pooled model--target contrasts favor SA, and each reported
target-level interval excludes zero. The smallest contrast is Gemma
on combinatorics/double counting at $+2.8$ points $[1.6, 4.1]$; the
largest is Qwen on combinatorics/invariant at $+27.9$ points
$[24.2, 31.8]$. The SA advantage therefore holds at every level of
disaggregation---across models, seeds, and individual target cells.

\subsection{The advantage is not only less negative transfer}

\begin{figure}[t]
\centering
\includegraphics[width=0.85\linewidth]{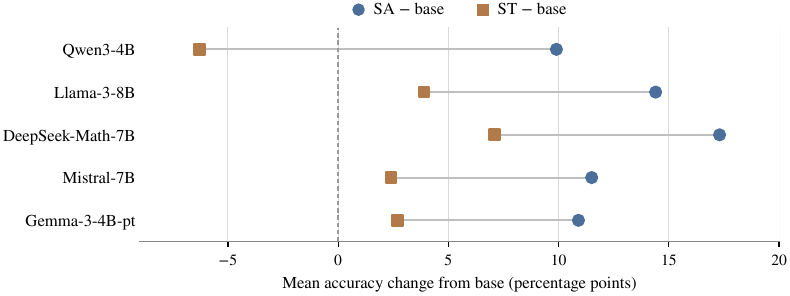}
\caption{\textbf{Positive and negative transfer coexist with the same
ordering.} Points are descriptive means of the four target-wise
changes, pooled over three seeds. SA improves average performance in
every model; ST improves it in four and lowers it in Qwen. Full
target-level CIs appear in Appendix~\ref{app:fullresults}.}
\label{fig:absolute}
\end{figure}

A natural objection is that SA might outperform ST simply because
topic-matched training is harmful. Figure~\ref{fig:absolute}
separates the SA--ST contrast from absolute changes relative to the
unadapted base model.

For Llama, DeepSeek, Mistral, and Gemma, both SA and ST improve
accuracy over the base in all four targets---but SA improves it
substantially more. Across these four models, SA gains average
roughly 11--17 percentage points while ST gains average 2--7 points.
The SA advantage therefore persists where adaptation is beneficial
under both relations, not only where ST degrades performance.

Qwen provides a complementary regime: SA averages $+9.9$ points over
base, while ST averages $-6.3$ points, with negative ST transfer in
every target. This demonstrates that topic matching does not guarantee
beneficial cross-approach adaptation. Across all 20 model--target
combinations, SA--base is positive in 19; the single exception is
Qwen on probability/double counting ($-0.3$ points $[-4.5, 3.8]$),
where SA still exceeds ST by $+6.3$ points.

Low base accuracies in some models should not be interpreted as
absence of pretrained mathematical knowledge, since base performance
depends on prompting format, generation strategy, and answer
extraction. The finding concerns improved target accuracy after
adaptation, not learning from a zero-knowledge starting point.

\subsection{Target-wise effects are consistent but heterogeneous}

\begin{figure}[t]
\centering
\includegraphics[width=0.7\linewidth]{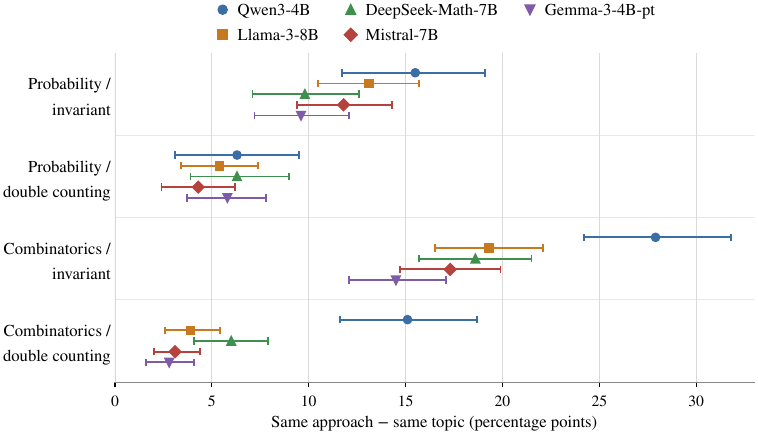}
\caption{\textbf{Every target favors SA.} All
points pool three seeds. Error bars are 95\% CIs. Invariant targets
show larger contrasts, particularly in combinatorics, but the ordering
is positive for both approaches in both topics. These are target-wise
effects, not estimates of an intrinsic transferability ranking among
approaches.}
\label{fig:targets}
\end{figure}

Figure~\ref{fig:targets} displays all 20 contrasts. Averaged across
models, the SA--ST gaps are $12.0$ points for probability/invariant,
$5.6$ for probability/double counting, $19.5$ for
combinatorics/invariant, and $6.2$ for combinatorics/double counting.
Combinatorics/invariant is the strongest target for every model, but
both double-counting targets retain positive contrasts throughout the
suite.

This heterogeneity does not undermine the aggregate finding but does
require careful interpretation. A globally stronger invariant source
would increase the invariant-target contrast (where it serves as SA)
and reduce the double-counting-target contrast (where it serves as
ST). Such a source effect cancels in the equally weighted aggregate
but not in each target subgroup. The target-wise contrasts therefore
do not distinguish source quality from a genuine target--approach
interaction, and we do not interpret the pattern as evidence that
invariant reasoning is intrinsically more transferable.

\subsection{The ordering replicates with different topics and approaches}
\label{sec:replication}

To test whether the observed ordering is specific to the topic and
approach choices in the primary design, we constructed a second
$2\times2$ dataset containing 800 problems. This design crosses number
theory and geometry with complement and pigeonhole reasoning. The dataset was constructed and validated using the same generation,
approach-verification, answer-auditing, and filtering pipeline as the
primary dataset. Each cell contains 200 problems, with 180 used for
training and 20 for validation; evaluation on a held-out target uses all
200 target problems. Training uses the same rank-16 LoRA protocol as the
primary experiment. We
repeated the rotational comparison across the same five base models
and training seeds 7, 17, and 27.

The model-level SA--ST advantages are $+16.0$ points for Llama,
$+15.7$ for Qwen, $+14.6$ for DeepSeek, $+13.0$ for Gemma, and
$+12.0$ for Mistral, averaging $+14.3$ percentage points. All five
model-level 95\% confidence intervals exclude zero. All 15 model--seed
aggregate contrasts and all 20 seed-pooled model--target point
estimates favor SA. SA also improves over the base model on average
in every model, with gains ranging from $+4.0$ to $+18.1$ points.
ST improves average accuracy in four models but reduces it in Qwen.

Statement-similarity diagnostics again favor ST: embedding
nearest-neighbor similarity is $0.727$ for ST versus $0.544$ for SA,
and TF--IDF nearest-neighbor similarity is $0.306$ versus $0.116$.
Thus, the SA-over-ST ordering replicates when both topics and
approaches change, despite greater measured statement resemblance
under ST. Appendix~\ref{app:replication} reports the model-level,
per-seed, and target-level results.

\subsection{The ordering persists under full fine-tuning}
\label{sec:fullft}

The primary experiments use rank-16 LoRA, raising the possibility that the SA advantage depends on the adaptation method. We therefore repeated the full rotation using full-parameter fine-tuning on all five models at seed 17.

Mistral did not converge under full fine-tuning, so we exclude it from the
summaries below and report it separately in Appendix~\ref{app:fullft}. For
the remaining four models, the SA--ST contrast is positive in all 16
model--target comparisons, with a mean advantage of $+11.1$ points versus
$+12.0$ in the matched LoRA runs. The SA-over-ST ordering is therefore not
specific to low-rank adaptation.

\subsection{The advantage appears at smaller training budgets}
\label{sec:size}
The contrasts above use 450 source-training examples. A budget ablation subsamples each source cell into nested, level-balanced subsets of 150, 300, and 450 problems and repeats the full rotation on all five models. It rests on weaker evidence than Section~\ref{sec:results} because it
uses a single seed. We therefore read it primarily for directional
robustness rather than precise effect-size estimation.

\begin{figure}[t]
\centering
\includegraphics[width=0.85\linewidth]{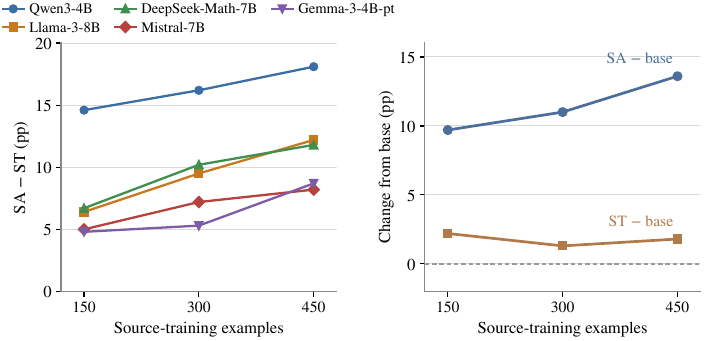}
\caption{\textbf{The ordering holds at every training budget and widens with data.} Left: SA--ST for each model at 150, 300, and 450 source-training examples. Right: suite-mean change from base for each relation. All five models; single seed; epochs are fixed, so optimizer steps grow with the data.}
\label{fig:size}
\end{figure}

Every model aggregate favors SA at 150, 300, and 450 exmaples. All five models increase monotonically, lifting the suite mean from $+7.5$ to $+9.7$ to $+11.4$  points across the three budgets (Figure~\ref{fig:size}, left). 

Separating the two relations shows where the widening originates (Figure~\ref{fig:size}, right). Averaged over the suite, SA--base grows from $+9.7$ to $+13.3$ points, while ST--base
stays small and relatively flat across budgets
($+2.2$, $+1.3$, $+1.9$). Additional same-topic supervision does not convert into target accuracy at these budgets, whereas additional same-approach supervision does. Appendix~\ref{app:size} reports the per-model values.

\section{Statement similarity and the transfer ordering}
\label{sec:similarity}

If the SA advantage were driven by greater surface resemblance between
source and target problems, we would expect SA sources to be more
similar to targets than ST sources. The opposite holds. Using
Qwen3-Embedding representations \citep{zhang2025embedding} and
TF--IDF, we compare source--target statement similarity under both
all-pairs averages and nearest-neighbor comparisons. For normalized
embeddings $e(x)$, the directional nearest-neighbor similarity is
\begin{equation}
 \mathrm{NN}(T,S)=\frac{1}{|T|}\sum_{x\in T}
 \max_{z\in S}\langle e(x),e(z)\rangle.
\label{eq:similarity}
\end{equation}

Across cells, statement embedding similarity favors ST over SA
for all pairs ($0.458$ versus $0.437$) and nearest neighbors
($0.739$ versus $0.705$). Statement TF--IDF nearest-neighbor
similarity likewise favors ST ($0.260$ versus $0.232$). For
probability/invariant, the nearest out-of-cell statement is in
the ST cell for $63.0\%$ of items and in the SA cell for
$35.4\%$. The model is evaluated on statements, so these are
the relevant resemblance cues. They favor ST; transfer favors
SA.
\begin{figure}[t]
\centering
\includegraphics[width=0.85\linewidth]{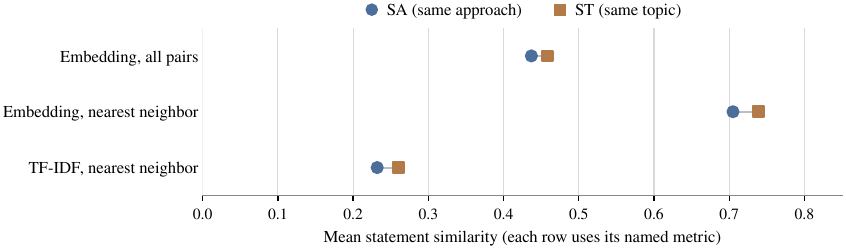}
\caption{\textbf{Greater statement resemblance does not track
greater transfer.} ST exceeds SA on every reported statement
measure. NN denotes nearest neighbor.}
\label{fig:similarity}
\end{figure}

\section{Discussion and limitations}
\label{sec:discussion}

The central finding is that approach-matched supervision consistently
outperforms topic-matched supervision in cross-cell mathematical
transfer. The result holds across five base models, three training seeds, and all
target rotations, yielding positive SA--ST contrasts in all 40
seed-pooled model--target comparisons. Because the second design changes
both the mathematical topics and reasoning approaches, the ordering is
not tied to a single factorial construction.

The designs were built as controlled tests in which topic sharing and
approach sharing can be varied independently while remaining
mathematically meaningful. This makes the repeated SA-over-ST ordering
across four topics and four approaches stronger evidence than a result
from one topic pair alone. The similarity diagnostics reinforce this
interpretation: in both designs, ST sources are more similar to targets
under lexical and embedding measures, yet they transfer less effectively.

The practical implication is that mathematical fine-tuning data should
not be organized by topic alone; reasoning approach should be considered
explicitly when selecting supervision for transfer. The study is still
a controlled investigation rather than an exhaustive survey of
mathematics, and broader domains and solution structures would provide
additional tests of generality. The synthetic construction also does not
eliminate every possible source--target interaction, and final-answer
accuracy does not identify the internal mechanism of transfer.


\section{Conclusion}
\label{sec:conclusion}

Across two independently constructed $2\times2$ designs, five base
models, and three training seeds per design, approach-matched
supervision outperforms topic-matched supervision in every one of the
40 seed-pooled model--target comparisons. The advantage is large and
replicable: it averages 10.8 percentage points in the first design and
14.3 points in the second, despite changing both the mathematical
topics and the reasoning approaches. The ordering also holds across
all model--seed aggregates and every target cell in both designs.

This advantage is not explained by greater surface resemblance.
Across both datasets, same-topic sources are more similar to their
targets under embedding and lexical measures, yet they transfer less
effectively. Thus, mathematical transfer follows shared reasoning
approach more consistently than shared topic in our experiments.

Taken together, these results provide evidence that reasoning approach
is a stronger organizing principle than topic for selecting
transfer-oriented mathematical supervision in LLMs. Rather than
treating mathematical domain as the primary criterion for choosing
training examples, our findings suggest that the structure of the
solution method should play a central role in how mathematical
fine-tuning data are organized and selected.

\clearpage
\section*{AI use statement}
AI systems were used to generate mathematical problems and reference solutions, produce unconstrained re-solutions, judge approach labels, and audit final answers. AI assisted with manuscript rewriting, literature research, reference checking, and preparation of LaTeX and figure code.

\section*{Reproducibility statement}
Section~\ref{sec:design} defines the estimand and rotation;
Appendices~\ref{app:fullresults} and~\ref{app:replication}
provide the complete summary statistics for the two designs.

\begingroup\small
\bibliography{iclr2027_conference}
\bibliographystyle{iclr2027_conference}
\endgroup
\clearpage
\appendix
\section{Complete results}
\label{app:fullresults}
Tables use seeds 7, 17, and 27. Intervals are those supplied with the summaries. Point estimates and differences were rounded separately, so a displayed subtraction can differ by 0.1.

\begin{table}[h]
\centering\small
\caption{Model-level SA--ST contrasts, pooled over four targets and three seeds.}
\label{tab:aggregate}
\begin{tabular}{@{}lrr@{}}\toprule
Model & SA--ST (pp) & 95\% CI \\\midrule
Qwen3-4B-Base & +16.2 & $[+14.4,+17.9]$ \\
Meta-Llama-3-8B & +10.5 & $[+9.4,+11.6]$ \\
DeepSeek-Math-7B & +10.2 & $[+8.9,+11.5]$ \\
Mistral-7B-v0.3 & +9.1 & $[+8.0,+10.2]$ \\
Gemma-3-4B-pt & +8.2 & $[+7.1,+9.3]$ \\
\midrule
Equal-weight model mean & +10.8 & --- \\
\bottomrule\end{tabular}
\end{table}

\begin{table}[h]
\centering\small
\caption{Per-seed SA--ST, averaged over the four targets.}
\label{tab:seeds}
\begin{tabular}{@{}lrrr@{}}\toprule
Model & Seed 7 & Seed 17 & Seed 27 \\\midrule
Qwen3-4B-Base & +15.8 & +16.5 & +16.2 \\
Meta-Llama-3-8B & +10.1 & +11.4 & +9.9 \\
DeepSeek-Math-7B & +11.4 & +10.4 & +8.8 \\
Mistral-7B-v0.3 & +9.5 & +9.1 & +8.8 \\
Gemma-3-4B-pt & +7.4 & +9.6 & +7.5 \\
\bottomrule\end{tabular}
\end{table}

\begin{table}[h]
\centering\small
\caption{Absolute accuracy (\%), pooled over the three seeds. P = probability; C = combinatorics; I = invariant; D = double counting.}
\label{tab:accuracy}
\begin{tabular}{@{}llrrr@{}}\toprule
Model & Target & Base & SA & ST \\\midrule
Qwen3-4B & P / I & 28.4 & 37.3 & 21.9 \\
 & P / D & 45.2 & 44.9 & 38.5 \\
 & C / I & 28.8 & 50.8 & 22.9 \\
 & C / D & 28.0 & 37.1 & 22.0 \\
\addlinespace[3pt]
Llama-3-8B & P / I & 2.2 & 17.9 & 4.8 \\
 & P / D & 0.4 & 12.2 & 6.8 \\
 & C / I & 0.8 & 25.7 & 6.4 \\
 & C / D & 0.4 & 5.5 & 1.6 \\
\addlinespace[3pt]
DeepSeek-Math-7B & P / I & 3.8 & 18.3 & 8.5 \\
 & P / D & 4.0 & 22.1 & 15.8 \\
 & C / I & 0.0 & 26.9 & 8.3 \\
 & C / D & 0.0 & 9.7 & 3.7 \\
\addlinespace[3pt]
Mistral-7B & P / I & 3.4 & 16.1 & 4.3 \\
 & P / D & 1.4 & 9.7 & 5.4 \\
 & C / I & 1.4 & 23.0 & 5.7 \\
 & C / D & 0.6 & 4.0 & 0.9 \\
\addlinespace[3pt]
Gemma-3-4B-pt & P / I & 1.8 & 14.1 & 4.5 \\
 & P / D & 0.8 & 10.3 & 4.5 \\
 & C / I & 0.4 & 18.8 & 4.3 \\
 & C / D & 0.2 & 3.7 & 0.9 \\
\bottomrule\end{tabular}
\end{table}

\clearpage
\begin{table}[t]
\centering\small
\caption{Target-level accuracy differences (percentage points) with 95\% CIs, pooled over seeds 7, 17, and 27.}
\label{tab:allci}
\setlength{\tabcolsep}{4pt}
\begin{tabular}{@{}lrrr@{}}\toprule
Target & SA--base [95\% CI] & ST--base [95\% CI] & SA--ST [95\% CI] \\\midrule
\multicolumn{4}{@{}l}{\textbf{Qwen3-4B-Base}} \\
P / I & $+8.9\;[+4.9,+13.0]$ & $-6.5\;[-10.5,-2.5]$ & $+15.5\;[+11.7,+19.1]$ \\
P / D & $-0.3\;[-4.5,+3.8]$ & $-6.7\;[-10.5,-2.7]$ & $+6.3\;[+3.1,+9.5]$ \\
C / I & $+22.0\;[+17.1,+27.0]$ & $-5.9\;[-10.0,-1.9]$ & $+27.9\;[+24.2,+31.8]$ \\
C / D & $+9.1\;[+4.7,+13.2]$ & $-6.0\;[-9.8,-2.3]$ & $+15.1\;[+11.6,+18.7]$ \\
\addlinespace[5pt]
\multicolumn{4}{@{}l}{\textbf{Meta-Llama-3-8B}} \\
P / I & $+15.7\;[+13.1,+18.3]$ & $+2.6\;[+0.7,+4.3]$ & $+13.1\;[+10.5,+15.7]$ \\
P / D & $+11.8\;[+9.8,+13.9]$ & $+6.4\;[+4.9,+8.0]$ & $+5.4\;[+3.4,+7.4]$ \\
C / I & $+24.9\;[+22.0,+27.8]$ & $+5.6\;[+4.1,+7.2]$ & $+19.3\;[+16.5,+22.1]$ \\
C / D & $+5.1\;[+3.7,+6.7]$ & $+1.2\;[+0.3,+2.0]$ & $+3.9\;[+2.6,+5.4]$ \\
\addlinespace[5pt]
\multicolumn{4}{@{}l}{\textbf{DeepSeek-Math-7B}} \\
P / I & $+14.5\;[+11.5,+17.6]$ & $+4.7\;[+2.3,+7.1]$ & $+9.8\;[+7.1,+12.6]$ \\
P / D & $+18.1\;[+14.9,+21.5]$ & $+11.8\;[+9.1,+14.5]$ & $+6.3\;[+3.9,+9.0]$ \\
C / I & $+26.9\;[+24.0,+29.7]$ & $+8.3\;[+6.6,+9.9]$ & $+18.6\;[+15.7,+21.5]$ \\
C / D & $+9.7\;[+7.9,+11.5]$ & $+3.7\;[+2.7,+4.7]$ & $+6.0\;[+4.1,+7.9]$ \\
\addlinespace[5pt]
\multicolumn{4}{@{}l}{\textbf{Mistral-7B-v0.3}} \\
P / I & $+12.7\;[+10.1,+15.3]$ & $+0.9\;[-1.1,+2.7]$ & $+11.8\;[+9.4,+14.3]$ \\
P / D & $+8.3\;[+6.2,+10.3]$ & $+4.0\;[+2.4,+5.6]$ & $+4.3\;[+2.4,+6.2]$ \\
C / I & $+21.6\;[+18.8,+24.5]$ & $+4.3\;[+2.7,+5.9]$ & $+17.3\;[+14.7,+19.9]$ \\
C / D & $+3.4\;[+2.0,+4.8]$ & $+0.3\;[-0.5,+1.0]$ & $+3.1\;[+2.0,+4.4]$ \\
\addlinespace[5pt]
\multicolumn{4}{@{}l}{\textbf{Gemma-3-4B-pt}} \\
P / I & $+12.3\;[+9.9,+14.7]$ & $+2.7\;[+1.1,+4.1]$ & $+9.6\;[+7.2,+12.1]$ \\
P / D & $+9.5\;[+7.4,+11.5]$ & $+3.7\;[+2.4,+4.9]$ & $+5.8\;[+3.7,+7.8]$ \\
C / I & $+18.4\;[+15.9,+21.0]$ & $+3.9\;[+2.7,+5.1]$ & $+14.5\;[+12.1,+17.1]$ \\
C / D & $+3.5\;[+2.3,+4.8]$ & $+0.7\;[+0.1,+1.4]$ & $+2.8\;[+1.6,+4.1]$ \\
\bottomrule\end{tabular}
\end{table}

\clearpage
\section{Second-design replication}
\label{app:replication}

The second dataset contains 800 problems and crosses number theory and
geometry with complement and pigeonhole reasoning. It was constructed
using the same generation, unconstrained re-solution, approach-label
verification, filtering, and final-answer auditing framework as the
primary dataset. The only changes are the topic--approach definitions
and dataset size.

Each of the four cells contains 200 problems. For a source cell, 180
problems are used for training and 20 for validation. Evaluation on a
held-out target cell uses all 200 target problems. Training uses the same
rank-16 LoRA configuration as the primary experiment, with seeds 7, 17,
and 27. The same rotational source--target assignment defined in
Section~\ref{sec:design} is applied to all four cells.

Table~\ref{tab:replication_models} reports model-level contrasts
averaged equally over the four target cells and three seeds.
Table~\ref{tab:replication_seeds} reports the individual seed-level
contrasts, and Table~\ref{tab:replication_cells} reports seed-pooled
target accuracies and contrasts. All 15 model--seed aggregate
contrasts and all 20 seed-pooled model--target point estimates
favor SA. Confidence intervals in
Table~\ref{tab:replication_models} are model-level intervals;
target-level intervals are not reported in these tables.
Point estimates and differences are rounded separately.

\begin{table}[htbp]
\centering
\small
\setlength{\tabcolsep}{4pt}
\caption{Second-design model-level accuracy differences, in
percentage points, averaged over four target cells and three seeds.
Confidence intervals are for SA--ST.}
\label{tab:replication_models}
\begin{tabular}{@{}lrrrr@{}}
\toprule
Model & SA--ST & 95\% CI & SA--base & ST--base \\
\midrule
Meta-Llama-3-8B
  & $+16.0$ & $[+14.0,+17.8]$ & $+18.1$ & $+2.1$ \\
Qwen3-4B-Base
  & $+15.7$ & $[+13.0,+18.3]$ & $+4.0$ & $-11.7$ \\
DeepSeek-Math-7B-Base
  & $+14.6$ & $[+12.6,+16.7]$ & $+17.7$ & $+3.1$ \\
Gemma-3-4B-pt
  & $+13.0$ & $[+11.2,+14.8]$ & $+14.9$ & $+1.9$ \\
Mistral-7B-v0.3
  & $+12.0$ & $[+10.2,+13.7]$ & $+13.1$ & $+1.1$ \\
\bottomrule
\end{tabular}
\end{table}

\begin{table}[htbp]
\centering
\small
\caption{Second-design SA--ST contrasts by training seed,
averaged over the four target cells. Values are in percentage points.}
\label{tab:replication_seeds}
\begin{tabular}{@{}lrrr@{}}
\toprule
Model & Seed 7 & Seed 17 & Seed 27 \\
\midrule
Meta-Llama-3-8B       & $+17.1$ & $+16.1$ & $+14.6$ \\
Qwen3-4B-Base         & $+14.2$ & $+16.2$ & $+16.5$ \\
DeepSeek-Math-7B-Base & $+15.1$ & $+13.4$ & $+15.2$ \\
Gemma-3-4B-pt         & $+15.1$ & $+10.6$ & $+13.1$ \\
Mistral-7B-v0.3       & $+10.0$ & $+11.9$ & $+14.1$ \\
\bottomrule
\end{tabular}
\end{table}

\clearpage
\begin{table}[t]
\centering
\small
\setlength{\tabcolsep}{4pt}
\caption{Second-design target-level accuracy (\%) and SA--ST
contrasts (percentage points), pooled over seeds 7, 17, and 27.
NT = number theory; Geo = geometry; Comp = complement;
Pig = pigeonhole. Differences are computed before rounding.}
\label{tab:replication_cells}
\begin{tabular}{@{}llrrrr@{}}
\toprule
Model & Target & Base & SA & ST & SA--ST \\
\midrule
Meta-Llama-3-8B
  & NT / Comp  & 1.0 & 5.7  & 0.7 & $+5.0$ \\
  & NT / Pig   & 0.0 & 25.5 & 4.5 & $+21.0$ \\
  & Geo / Comp & 0.0 & 3.8  & 1.5 & $+2.3$ \\
  & Geo / Pig  & 0.5 & 38.8 & 3.3 & $+35.5$ \\
\addlinespace[3pt]
Qwen3-4B-Base
  & NT / Comp  & 37.5 & 41.3 & 38.0 & $+3.3$ \\
  & NT / Pig   & 49.0 & 53.2 & 29.2 & $+24.0$ \\
  & Geo / Comp & 27.0 & 38.5 & 28.3 & $+10.2$ \\
  & Geo / Pig  & 65.5 & 61.8 & 36.7 & $+25.2$ \\
\addlinespace[3pt]
DeepSeek-Math-7B-Base
  & NT / Comp  & 0.0 & 7.3  & 2.7 & $+4.7$ \\
  & NT / Pig   & 1.0 & 25.2 & 4.2 & $+21.0$ \\
  & Geo / Comp & 1.0 & 6.3  & 4.0 & $+2.3$ \\
  & Geo / Pig  & 0.5 & 34.3 & 4.0 & $+30.3$ \\
\addlinespace[3pt]
Gemma-3-4B-pt
  & NT / Comp  & 0.0 & 4.3  & 1.5 & $+2.8$ \\
  & NT / Pig   & 0.0 & 23.3 & 1.5 & $+21.8$ \\
  & Geo / Comp & 0.5 & 4.2  & 1.5 & $+2.7$ \\
  & Geo / Pig  & 0.0 & 28.2 & 3.7 & $+24.5$ \\
\addlinespace[3pt]
Mistral-7B-v0.3
  & NT / Comp  & 0.5 & 2.3  & 1.5 & $+0.8$ \\
  & NT / Pig   & 1.5 & 18.7 & 3.3 & $+15.3$ \\
  & Geo / Comp & 0.5 & 3.7  & 0.5 & $+3.2$ \\
  & Geo / Pig  & 1.0 & 31.2 & 2.5 & $+28.7$ \\
\bottomrule
\end{tabular}
\end{table}

\clearpage
\section{Full fine-tuning ablation}
\label{app:fullft}
This check updates all language-model weights instead of rank-16 LoRA adapters, on the same five models, at seed 17. Settings otherwise follow the LoRA runs (three epochs, effective batch 16, validation-loss checkpointing) at lr $10^{-5}$. Models up to 4B use AdamW; the 7--8B models use Adafactor. 

\begin{table}[h]
\centering\small
\caption{Full fine-tuning, averaged over the four target cells. The LoRA column is the matched low-rank run. $^{\dagger}$Mistral did not converge and is excluded from the mean.}
\label{tab:fullft}
\setlength{\tabcolsep}{5pt}
\begin{tabular}{@{}lrrrr@{}}\toprule
Model & SA--ST & SA--base & ST--base & LoRA SA--ST \\\midrule
Qwen3-4B-Base & +19.8 & +11.0 & $-8.8$ & +16.5 \\
Meta-Llama-3-8B & +6.9 & +7.5 & +0.6 & +11.4 \\
DeepSeek-Math-7B-Base & +12.2 & +19.8 & +7.6 & +10.4 \\
Gemma-3-4B-pt & +5.4 & +8.0 & +2.6 & +9.6 \\
\midrule
Mean of the four above & +11.1 & --- & --- & +12 \\
\midrule
Mistral-7B-v0.3$^{\dagger}$ & +0.5 & +0.4 & $-0.1$ & +9.1 \\
\bottomrule\end{tabular}
\end{table}

\begin{table}[h]
\centering\small
\caption{Target-level SA--ST contrasts under full fine-tuning (percentage points).}
\label{tab:fullfttargets}
\begin{tabular}{@{}lrrrr@{}}\toprule
Model & P / I & P / D & C / I & C / D \\\midrule
Qwen3-4B-Base & +28.0 & +13.0 & +25.0 & +13.0 \\
Meta-Llama-3-8B & +8.2 & +1.0 & +16.0 & +2.4 \\
DeepSeek-Math-7B-Base & +10.6 & +8.8 & +23.6 & +5.6 \\
Gemma-3-4B-pt & +10.4 & +0.4 & +10.4 & +0.4 \\
Mistral-7B-v0.3$^{\dagger}$ & +0.4 & +0.8 & +0.4 & +0.4 \\
\bottomrule\end{tabular}
\end{table}

Mistral's four contrasts are $+0.4$ to $+0.8$ points and its SA accuracies fall below base on two targets; the row is reported for completeness. Excluding it, the mean is $+11.1$ against $+12.0$ for the matched LoRA runs.

\clearpage
\section{Training-set size ablation}
\label{app:size}
Each source cell is subsampled into nested, level-balanced subsets ($150\subset300\subset450$), with the same 50 validation items and evaluation on all 500 target items. Every size trains for three epochs, so the step budget grows with the data (about 30, 57, and 87 steps).

\begin{table}[h]
\centering\small
\caption{Accuracy differences (percentage points) by number of source-training examples, averaged over the four target cells.}
\label{tab:size}
\setlength{\tabcolsep}{3.5pt}
\begin{tabular}{@{}lrrrrrrrrr@{}}\toprule
& \multicolumn{3}{c}{SA--ST} & \multicolumn{3}{c}{SA--base} & \multicolumn{3}{c}{ST--base} \\
\cmidrule(lr){2-4}\cmidrule(lr){5-7}\cmidrule(lr){8-10}
Model & 150 & 300 & 450 & 150 & 300 & 450 & 150 & 300 & 450 \\\midrule
Qwen3-4B-Base & +14.6 & +16.2 & +16.5 & +11.1 & +9.5 & +10.2 & $-3.5$ & $-6.7$ & $-6.3$ \\
Meta-Llama-3-8B & +6.4 & +9.5 & +11.4 & +10.5 & +12.7 & +14.9 & +4.1 & +3.1 & +3.5 \\
DeepSeek-Math-7B-Base & +6.7 & +10.2 & +10.4  & +12.9 & +15.7 & +17.7 & +6.2 & +5.5 & +7.3 \\
Mistral-7B-v0.3 & +5.0 & +7.2 & +9.1  & +7.0 & +8.9 & +11.7 & +2.0 & +1.8 & +2.6\\
Gemma-3-4B-pt & +4.8 & +5.3 & +9.6 & +7.1 & +8.0 & +11.9 & +2.2 & +2.7 & +2.3 \\
\midrule
Suite mean & +7.5 & +9.7 & +11.4 & +9.7 & +11.0 & +13.3 & +2.2 & +1.3 & +1.9 \\
\bottomrule\end{tabular}
\end{table}

\clearpage
\section{Checkpoints and protocol}
\label{app:protocol}

\begin{table}[h]
\centering\small
\caption{Public checkpoints used in the primary suite.}
\begin{tabular}{@{}ll@{}}
\toprule
Model & Checkpoint \\
\midrule
Qwen3-4B-Base &
\texttt{Qwen/Qwen3-4B-Base} \\
Meta-Llama-3-8B &
\texttt{meta-llama/Meta-Llama-3-8B} \\
DeepSeek-Math-7B-Base &
\texttt{deepseek-ai/deepseek-math-7b-base} \\
Mistral-7B-v0.3 &
\texttt{mistralai/Mistral-7B-v0.3} \\
Gemma-3-4B-pt &
\texttt{google/gemma-3-4b-pt} \\
\bottomrule
\end{tabular}
\end{table}

Each cell contains 500 problems. For a source cell, 450 problems are
used for training and the remaining 50 for validation. Evaluation on
a held-out target cell uses all 500 target problems. The primary
setting uses rank-16 LoRA with training seeds 7, 17, and 27, yielding
$5\times3\times4=60$ source-training jobs.

Approach labels correspond to the two construction categories defined
in Section~\ref{sec:design}. Invariant reasoning uses a quantity or
relation preserved under the transformations in the problem, whereas
double counting derives a result by counting the same collection or
incidence structure in two ways. A problem was retained only when all
three approach judges agreed on its label.

\section{Statistical analysis}
\label{app:statistics}

\paragraph{Confidence intervals.}
We use a paired, cell-stratified nonparametric percentile bootstrap
over evaluation problems. Within each target cell, evaluation problems are resampled with replacement, preserving the original number of evaluation problems
in that cell. For the SA--ST contrast, let

\[
d_i =
y_i^{\mathrm{SA}} -
y_i^{\mathrm{ST}},
\]

where $y_i^{\mathrm{SA}}$ and $y_i^{\mathrm{ST}}$ are the binary
strict-correct outcomes for problem $i$. For each bootstrap replicate, we resample the evaluation problems
within that target cell with replacement and compute the mean of $d_i$. The 95\% confidence interval is given by the 2.5th and 97.5th
percentiles of the resulting bootstrap distribution. SA--base and
ST--base intervals are computed analogously. Because SA, ST, and base
are evaluated on the same target problems, outcomes remain paired
within each resampled problem.

For quantities averaged across the four target cells, each target cell
is resampled independently within its own evaluation set. The four
cell-level means are then equally averaged within each bootstrap
replicate, and the confidence interval is obtained from the
percentiles of this replicate-level distribution.

For results pooled across seeds 7, 17, and 27, each problem's binary
correctness outcome is first averaged across seeds. The resulting
per-problem outcome therefore takes a value in
$\{0,\frac{1}{3},\frac{2}{3},1\}$. Bootstrap resampling is then
performed over these seed-averaged problem-level outcomes using the
same paired, cell-stratified procedure.

The primary analyses use 4{,}000 bootstrap replicates. Bootstrap
sampling is initialized with
\texttt{np.random.default\_rng(7)} for reproducibility. Reported
confidence intervals quantify uncertainty arising from the finite set
of evaluation problems conditional on the trained checkpoints.
Training runs themselves are not resampled, and seeds are averaged
rather than resampled; consequently, the intervals do not represent
training-seed uncertainty. We report the individual seed-level
SA--ST contrasts separately to characterize variation across training
seeds.

Reported 95\% confidence intervals are pointwise and are not adjusted
for multiple comparisons.

\paragraph{Base predictions across seeds.}
Base-model predictions are reused across training seeds because base
evaluation uses deterministic greedy decoding and does not depend on
the fine-tuning seed.

\clearpage
\section{Similarity diagnostics}
\label{app:similarity}
\begin{table}[h]
\centering\small
\caption{Statement similarity by relation. Centred embeddings subtract the dataset-mean vector. Metrics are not interchangeable.}
\begin{tabular}{@{}lrrr@{}}\toprule
Metric & Within & SA & ST \\\midrule
TF--IDF, all pairs & 0.054 & 0.023 & 0.026 \\
TF--IDF, NN & 0.538 & 0.232 & 0.260 \\
Embedding, all pairs & 0.537 & 0.437 & 0.458 \\
Embedding, NN & 0.898 & 0.705 & 0.739 \\
Centred emb., all pairs & 0.153 & $-0.031$ & 0.004 \\
Centred emb., NN & 0.811 & 0.445 & 0.518 \\
\bottomrule\end{tabular}
\end{table}

\begin{table}[h]
\centering\small
\caption{Share of problems whose nearest out-of-cell statement neighbor lies in the SA or ST cell (Qwen3 embeddings). The leftover share matches neither topic nor approach.}
\begin{tabular}{@{}lrr@{}}\toprule
Cell & SA & ST \\\midrule
P / I & 35.4\% & 63.0\% \\
C / I & 52.8\% & 47.0\% \\
P / D & 45.8\% & 54.2\% \\
C / D & 33.2\% & 64.4\% \\
\bottomrule\end{tabular}
\end{table}

\end{document}